# Probabilistic Latent Semantic Analysis

Thomas Hofmann
EECS Department, Computer Science Division, University of California, Berkeley &
International Computer Science Institute, Berkeley, CA
hofmann@cs.berkeley.edu

## Abstract

*Probabilistic Latent Semantic Analysis* is a novel statistical technique for the analysis of two-mode and co-occurrence data, which has applications in information retrieval and filtering, natural language processing, machine learning from text, and in related areas. Compared to standard Latent Semantic Analysis which stems from linear algebra and performs a Singular Value Decomposition of co-occurrence tables, the proposed method is based on a mixture decomposition derived from a latent class model. This results in a more principled approach which has a solid foundation in statistics. In order to avoid overfitting, we propose a widely applicable generalization of maximum likelihood model fitting by tempered EM. Our approach yields substantial and consistent improvements over Latent Semantic Analysis in a number of experiments.

## 1 Introduction

Learning from text and natural language is one of the great challenges of Artificial Intelligence and Machine Learning. Any substantial progress in this domain has strong impact on many applications ranging from information retrieval, information filtering, and intelligent interfaces, to speech recognition, natural language processing, and machine translation. One of the fundamental problems is to learn the *meaning* and *usage* of words in a data-driven fashion, i.e., from some given text corpus, possibly without further linguistic prior knowledge.

The main challenge a machine learning system has to address roots in the distinction between the lexical level of "what actually has been said or written" and the semantical level of "what was intended" or "what was referred to" in a text or an utterance. The resulting problems are twofold: (i) polysems, i.e., a word may have multiple senses and multiple types of usage in different context, and (ii) synonymys and semantically related words, i.e., different words may have a similar meaning, they may at least in certain contexts denote the same concept or – in a weaker sense – refer to the same topic.

*Latent semantic analysis* (LSA) [3] is well-known technique which partially addresses these questions. The key idea is to map high-dimensional count vectors, such as the ones arising in vector space representations of text documents [12], to a lower dimensional representation in a so-called *latent semantic space*. As the name suggests, the goal of LSA is to find a data mapping which provides information well beyond the lexical level and reveals semantical relations between the entities of interest. Due to its generality, LSA has proven to be a valuable analysis tool with a wide range of applications (e.g. [3, 5, 8, 1]). Yet its theoretical foundation remains to a large extent unsatisfactory and incomplete.

This paper presents a statistical view on LSA which leads to a new model called *Probabilistic Latent Semantics Analysis* (PLSA). In contrast to standard LSA, its probabilistic variant has a sound statistical foundation and defines a proper generative model of the data. A detailed discussion of the numerous advantages of PLSA can be found in subsequent sections.

## 2 Latent Semantic Analysis

### 2.1 Count Data and Co-occurrence Tables

LSA can in principle be applied to any type of count data over a discrete dyadic domain (cf. [7]). However, since the most prominent application of LSA is in the analysis and retrieval of text documents, we focus on this setting for sake of concreteness. Suppose therefore we have given a collection of text doc-uments $\mathcal{D} = \{d_1, \ldots, d_N\}$ with terms from a vocabulary $\mathcal{W} = \{w_1, \ldots w_M\}$. By ignoring the sequential order in which words occur in a document, one may summarize the data in a $N \times M$ *co-occurrence table* of counts $\mathbf{N} = (n(d_i, w_j))_{ij}$, where $n(d, w) \in \mathbb{N}$ denotes how often the term $w$ occurred in document $d$. In this particular case, $\mathbf{N}$ is also called the term-document matrix and the rows/columns of $\mathbf{N}$ are referred to as document/term vectors, respectively. The key assumption is that the simplified 'bag-of-words' or vector-space representation [12] of documents will in many cases preserve most of the relevant information, e.g., for tasks like text retrieval based on keywords.

## 2.2 Latent Semantic Analysis by SVD

As mentioned in the introduction, the key idea of LSA is to map documents (and by symmetry terms) to a vector space of reduced dimensionality, the *latent semantic space* [3]. The mapping is restricted to be linear and is based on a Singular Value Decomposition (SVD) of the co-occurrence table. One thus starts with the standard SVD given by $\mathbf{N} = \mathbf{U\Sigma V}^t$, where $\mathbf{U}$ and $\mathbf{V}$ are orthogonal matrices $\mathbf{U}^t\mathbf{U} = \mathbf{V}^t\mathbf{V} = \mathbf{I}$ and the diagonal matrix $\mathbf{\Sigma}$ contains the singular values of $\mathbf{N}$. The LSA approximation of $\mathbf{N}$ is computed by setting all but the largest $K$ singular values in $\mathbf{\Sigma}$ to zero ($= \tilde{\mathbf{\Sigma}}$), which is rank $K$ optimal in the sense of the $L_2$-matrix norm. One obtains the approximation $\tilde{\mathbf{N}} = \mathbf{U}\tilde{\mathbf{\Sigma}}\mathbf{V}^t \approx \mathbf{U\Sigma V}^t = \mathbf{N}$. Notice that the document-to-document inner products based on this approximation are given by $\tilde{\mathbf{N}}\tilde{\mathbf{N}}^t = \mathbf{U}\tilde{\mathbf{\Sigma}}^2\mathbf{U}^t$ and hence one might think of the rows of $\mathbf{U}\tilde{\mathbf{\Sigma}}$ as defining coordinates for documents in the latent space. While the original high-dimensional vectors are sparse, the corresponding low-dimensional latent vectors will typically not be sparse. This implies that it is possible to compute meaningful association values between pairs of documents, even if the documents do not have any terms in common. The hope is that terms having a common meaning, in particular synonyms, are roughly mapped to the same direction in the latent space.

## 3 Probabilistic LSA

### 3.1 The Aspect Model

The starting point for *Probabilistic Latent Semantic Analysis* is a statistical model which has been called *aspect model* [7]. The aspect model is a latent variable model for co-occurrence data which associates an unobserved class variable $z \in \mathcal{Z} = \{z_1, \ldots, z_K\}$ with each observation. A joint probability model over $\mathcal{D} \times \mathcal{W}$ is

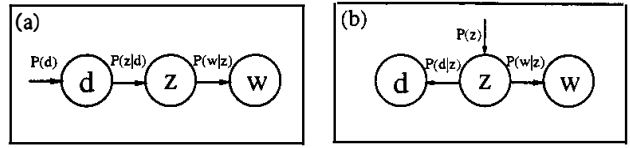

Figure 1: Graphical model representation of the aspect model in the asymmetric (a) and symmetric (b) parameterization.

defined by the mixture

$$P(d,w) = P(d)P(w|d), \quad P(w|d) = \sum_{z \in \mathcal{Z}} P(w|z)P(z|d). \quad (1)$$

Like virtually all statistical latent variable models the aspect model introduces a conditional independence assumption, namely that $d$ and $w$ are independent conditioned on the state of the associated latent variable (the corresponding graphical model representation is depicted in Figure 1 (a)). Since the cardinality of $z$ is smaller than the number of documents/words in the collection, $z$ acts as a bottleneck variable in predicting words. It is worth noticing that the model can be equivalently parameterized by (cf. Figure 1 (b))

$$P(d,w) = \sum_{z \in \mathcal{Z}} P(z)P(d|z)P(w|z), \quad (2)$$

which is perfectly symmetric in both entities, documents and words.

### 3.2 Model Fitting with the EM Algorithm

The standard procedure for maximum likelihood estimation in latent variable models is the Expectation Maximization (EM) algorithm [4]. EM alternates two coupled steps: (i) an expectation (E) step where posterior probabilities are computed for the latent variables, (ii) an maximization (M) step, where parameters are updated. Standard calculations (cf. [7, 13]) yield the E-step equation

$$P(z|d,w) = \frac{P(z)P(d|z)P(w|z)}{\sum_{z' \in \mathcal{Z}} P(z')P(d|z')P(w|z')}, \quad (3)$$

as well as the following M-step formulae

$$P(w|z) \propto \sum_{d \in \mathcal{D}} n(d,w)P(z|d,w), \quad (4)$$

$$P(d|z) \propto \sum_{w \in \mathcal{W}} n(d,w)P(z|d,w), \quad (5)$$

$$P(z) \propto \sum_{d \in \mathcal{D}} \sum_{w \in \mathcal{W}} n(d,w)P(z|d,w). \quad (6)$$

Before discussing further algorithmic refinements, we will study the relationship between the proposed model and LSA in more detail.

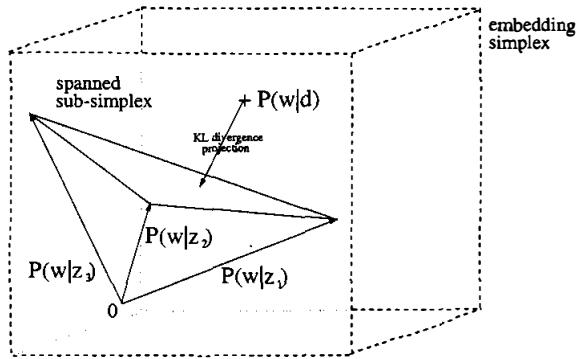

Figure 2: Sketch of the probability sub-simplex spanned by the aspect model.

### 3.3 Probabilistic Latent Semantic Space

Consider the class-conditional multinomial distributions $P(\cdot|z)$ over the vocabulary which we call *factors*. They can be represented as points on the $M-1$ dimensional simplex of all possible multinomials. Via its convex hull, this set of $K$ points defines a $L \leq K-1$ dimensional sub-simplex. The modeling assumption expressed by (1) is that conditional distributions $P(w|d)$ for all document are approximated by a multinomial representable as a convex combination of factors $P(w|z)$, where the mixing weights $P(z|d)$ uniquely define a point on the spanned sub-simplex. A simple sketch of this situation is shown in Figure 2. Despite of the discreteness of the introduced latent variables, a *continuous latent space* is obtained within the space of all multinomial distributions. Since the dimensionality of the sub-simplex is $\leq K-1$ as opposed to a maximum of $M-1$ for the complete probability simplex, this performs a dimensionality reduction in the space of multinomial distributions and the spanned sub-simplex can be identified with a *probabilistic latent semantic space*.

To stress this point and to clarify the relation to LSA, let us rewrite the aspect model as parameterized by (2) in matrix notation. Hence define matrices by $\hat{\mathbf{U}} = (P(d_i|z_k))_{i,k}$, $\hat{\mathbf{V}} = (P(w_j|z_k))_{j,k}$, and $\hat{\mathbf{\Sigma}} = \text{diag}(P(z_k))_k$. The joint probability model $\mathbf{P}$ can then be written as a matrix product $\mathbf{P} = \hat{\mathbf{U}}\hat{\mathbf{\Sigma}}\hat{\mathbf{V}}^t$. Comparing this with SVD, one can make the following observations: (i) outer products between rows of $\hat{\mathbf{U}}$ and $\hat{\mathbf{V}}$ reflect conditional independence in PLSA, (ii) the $K$ factors correspond to the mixture components in the aspect model, (iii) the mixing proportions in PLSA substitute the singular values. The crucial difference between PLSA and LSA, however, is the objective function utilized to determine the optimal decomposition/approximation. In LSA, this is the $L_2$- or Frobenius norm, which corresponds to an implicit additive Gaussian noise assumption on (possibly transformed) counts. In contrast, PLSA relies on the likelihood function of multinomial sampling and aims at an explicit maximization of the predictive power of the model. As is well known, this corresponds to a minimization of the cross entropy or Kullback-Leibler divergence between the empirical distribution and the model, which is very different from any type of squared deviation. On the modeling side this offers important advantages, for example, the mixture approximation $\mathbf{P}$ of the co-occurrence table is a well-defined probability distribution and factors have a clear probabilistic meaning. In contrast, LSA does not define a properly normalized probability distribution and $\tilde{\mathbf{N}}$ may even contain negative entries. In addition, there is no obvious interpretation of the directions in the LSA latent space, while the directions in the PLSA space are interpretable as multinomial word distributions. The probabilistic approach can also take advantage of the well-established statistical theory for model selection and complexity control, e.g., to determine the optimal number of latent space dimensions. Choosing the number of dimensions in LSA on the other hand is typically based on ad hoc heuristics.

A comparison of the computational complexity might suggest some advantages for LSA: ignoring potential problems of numerical stability the SVD can be computed exactly, while the EM algorithm is an iterative method which is only guaranteed to find a local maximum of the likelihood function. However, in all our experiments the computing time of EM has not been significantly worse than performing an SVD on the co-occurrence matrix. There is also a large potential for improving run-time performance of EM by on-line update schemes, which has not been explored so far.

### 3.4 Topic Decomposition and Polysemy

Let us briefly discuss some elucidating examples at this point which will also reveal a further advantage of PLSA over LSA in the context of polsemous words. We have generated a dataset (CLUSTER) with abstracts of 1568 documents on *clustering* and trained an aspect model with 128 latent classes. Four pairs of factors are visualized in Figure 3. These pairs have been selected as the two factors that have the highest probability to generate the words "segment", "matrix", "line", and "power", respectively. The sketchy characterization of the factors by their 10 most probable words already reveals interesting topics. In particular, notice that the term used to select a particular pair has a different meaning in either topic factor: (i) 'Segment' refers to an image region in the first and to a phonetic segment in the second factor. (ii) 'Matrix' denotes a rectangular table of numbers and to a material in which something is embedded or enclosed. (iii) 'Line' can refer to a line in an image, but also to a line in a spectrum.

| "segment 1" | "segment 2" | "matrix 1" | "matrix 2" | "line 1" | "line 2" | "power 1" | power 2" |
|---|---|---|---|---|---|---|---|
| imag | speaker | robust | manufactur | constraint | alpha | POWER | load |
| SEGMENT | speech | MATRIX | cell | LINE | redshift | spectrum | memori |
| texture | recogni | eigenvalu | part | match | LINE | omega | vlsi |
| color | signal | uncertainti | MATRIX | locat | galaxi | mpc | POWER |
| tissue | train | plane | cellular | imag | quasar | hsup | systolic |
| brain | hmm | linear | famili | geometr | absorp | larg | input |
| slice | source | condition | design | impos | high | redshift | complex |
| cluster | speakerind. | perturb | machinepart | segment | ssup | galaxi | arrai |
| mri | SEGMENT | root | format | fundament | densiti | standard | present |
| volume | sound | suffici | group | recogn | veloc | model | implement |

Figure 3: Eight selected factors from a 128 factor decomposition. The displayed word stems are the 10 most probable words in the class-conditional distribution $P(w|z)$, from top to bottom in descending order.

Document 1, $P\{z_k|d_1, w_j = \text{'segment'}\} = (0.951, 0.0001, \ldots)$
$P\{w_j = \text{'segment'}|d_1\} = 0.06$

SEGMENT medic imag challeng problem field imag analysi diagnost base proper SEGMENT digit imag SEGMENT medic imag need applic involv estim boundari object classif tissu abnorm shape analysi contour detec textur SEGMENT despit exist techniqu SEGMENT specif medic imag remain crucial problem [...]

Document 2, $P\{z_k|d_2, w_j = \text{'segment'}\} = (0.025, 0.867, \ldots)$
$P\{w_j = \text{'segment'}|d_2\} = 0.010$

consid signal origin sequenc sourc specif problem SEGMENT signal relat SEGMENT sourc address issu wide applic field report describ resolu method ergod hidden markov model hmm hmm state correspond signal sourc signal sourc sequenc determin decod procedur viterbi algorithm forward algorithm observ sequenc baumwelch train estim hmm paramet train materi applic multipl signal sourc identif problem experi perform unknown speaker identif [...]

Figure 4: Abstracts of 2 exemplary documents from the CLUSTER collection along with latent class posterior probabilities $P\{z|d, w = \text{'segment'}\}$ and word probabilities $P\{w = \text{'segment'}|d\}$.

(iv) 'Power' is used in the context of radiating objects in astronomy, but also in electrical engineering.

Figure 4 shows the abstracts of two exemplary documents which have been pre-processed by a standard stop-word list and a stemmer. The posterior probabilities for the classes given the different occurrences of 'segment' indicate how likely it is for each of the factors in the first pair of Figure 3 to have generated this observation. We have also displayed the estimates of the conditional word probabilities $P\{w = \text{'segment'}|d_{1,2}\}$. One can see that the correct meaning of the word 'segment' is identified in both cases. This implies that although 'segment' occurs frequently in both document, the overlap in the factored representation is low, since 'segement' is identified as a polysemous word (relative to the chosen resolution level) which – dependent on the context – is explained by different factors.

### 3.5 Aspects versus Clusters

It is worth comparing the aspect model with statistical clustering models (cf. also [7]). In clustering models for documents, one typically associates a latent class variable with each document in the collection. Most closely related to our approach is the *distributional clustering model* [10, 7] which can be thought of as an unsupervised version of a naive Bayes' classifier. It can be shown that the conditional word probability of a probabilistic clustering model is given by

$$P(w|d) = \sum_{z \in \mathcal{Z}} P\{c(d) = z\} P(w|z), \qquad (7)$$

where $P\{c(d) = z\}$ is the posterior probability of document $d$ having latent class $z$. It is a simple implication of Bayes' rule that these posterior probabilities will concentrate their probability mass on a certain value $z$ with an increasing number of observations (i.e., with the length of the document). This means that although (1) and (7) are algebraically equivalent, they are conceptually very different and yield in fact different results. The aspect model assumes that document-specific distributions are a convex combination of aspects, while the clustering model assumes there is just *one* cluster-specific distribution which is inherited by all documents in the cluster.[1] Thus in clustering models the class-conditionals $P(w|z)$ have to capture the complete vocabulary of a subset (cluster) of documents, while factors can focus on certain aspects of the vocabulary of a subset of documents. For example, a factor can be very well used to explain some *fraction* of the words occurring in a document, although it might not explain other words at all (e.g., even assign zero probability), because these other words can be taken care of by other factors.

---
[1] In the distributional clustering model it is only the posterior uncertainty of the cluster assignments that induces some averaging over the class-conditional word distributions $P(w|z)$.

### 3.6 Model Fitting Revisited: Improving Generalization by Tempered EM

So far we have focused on maximum likelihood estimation to fit a model from a given document collection. Although the likelihood or, equivalently, the perplexity[2] is the quantity we believe to be crucial in assessing the quality of a model, one clearly has to distinguish between the performance on the training data and on unseen test data. To derive conditions under which generalization on unseen data can be guaranteed is actually *the* fundamental problem of statistical learning theory. Here, we propose a generalization of maximum likelihood for mixture models which is known as *annealing* and is based on an entropic regularization term. The resulting method is called *Tempered Expectation Maximization* (TEM) and is closely related to *deterministic annealing* [11].

The starting point of TEM is a derivation of the E-step based on an optimization principle. As has been pointed out in [9] the EM procedure in latent variable models can be obtained by minimizing a common objective function – the (Helmholtz) *free energy* – which for the aspect model is given by

$$\mathcal{F}_\beta = -\beta \sum_{d,w} n(d,w) \sum_z \tilde{P}(z;d,w) \log P(d,w|z)P(z)$$
$$+ \sum_{d,w} n(d,w) \sum_z \tilde{P}(z;d,w) \log \tilde{P}(z;d,w). \quad (8)$$

Here $\tilde{P}(z;d,w)$ are variational parameters which define a conditional distribution over $\mathcal{Z}$ and $\beta$ is a parameter which – in analogy to physical systems – is called the *inverse computational temperature*. Notice that the first contribution in (8) is the negative expected log-likelihood scaled by $\beta$. Thus in the case of $\tilde{P}(z;d,w) = P(z|d,w)$ minimizing $\mathcal{F}$ w.r.t. the parameters defining $P(d,w|z)P(z)$ amounts to the standard M-step in EM. In fact, it is straightforward to verify that the posteriors are obtained by minimizing $\mathcal{F}$ w.r.t. $\tilde{P}$ at $\beta = 1$. In general $\tilde{P}$ is determined by

$$\tilde{P}(z;d,w) = \frac{[P(z)P(d|z)P(w|z)]^\beta}{\sum_{z'}[P(z')P(d|z')P(w|z')]^\beta}. \quad (9)$$

---

[2]Perplexity refers to the log-averaged inverse probability on unseen data.

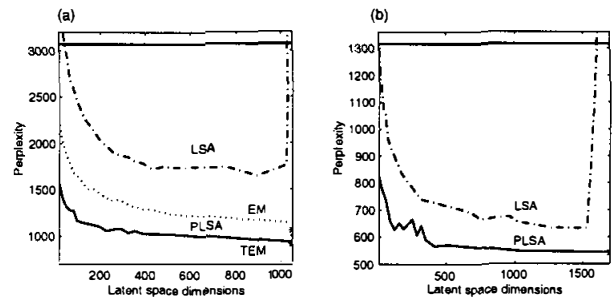

Figure 5: Perplexity results as a function of the latent space dimensionality for (a) the MED data (rank 1033) and (b) the LOB data (rank 1674). Plotted results are for LSA (dashed-dotted curve) and PLSA (trained by TEM = solid curve, trained by early stopping EM = dotted curve). The upper baseline is the unigram model corresponding to marginal independence. The star at the right end of the PLSA denotes the perplexity of the largest trained aspect models ($K = 2048$).

This shows that the effect of the entropy at $\beta < 1$ is to dampen the posterior probabilities such that they will become closer to the uniform distribution with decreasing $\beta$.

Somewhat contrary to the spirit of annealing as a continuation method, we propose an 'inverse' annealing strategy which first performs EM iterations and then *decreases* $\beta$ until performance on held-out data deteriorates. Compared to annealing this may accelerate the model fitting procedure significantly (e.g., by a factor of $\approx 10 - 50$) and we have not found the test set performance of 'heated' models to be worse than the one achieved by carefully 'annealed' models. The TEM algorithm can thus be implemented in the following way:

1. Set $\beta \leftarrow 1$ and perform EM with early stopping.

2. Decrease $\beta \leftarrow \eta\beta$ (with $\eta < 1$) and perform one TEM iteration.

3. As long as the performance on held-out data improves (non-negligible) continue TEM iterations at this value of $\beta$, otherwise goto step 2

4. Perform stopping on $\beta$, i.e., stop when decreasing $\beta$ does not yield further improvements.

## 4 Experimental Results

In the experimental evaluation, we focus on two tasks: (i) perplexity minimization for a document-specific unigram model and noun-adjective pairs, and (ii) automated indexing of documents. The evaluation of LSA and PLSA on the first task will demonstrate the advantages of explicitly minimizing perplexity by TEM, the second task will show that the solid statistical foundation of PLSA pays off even in applications which are not directly related to perplexity reduction.

Table 1: Average precision results and relative improvement w.r.t. the baseline method cos+tf for the 4 standard test collections. Compared are LSI, PLSI, as well as results obtained by combining PLSI models (PLSI*). An asterix for LSI indicates that no performance gain could be achieved over the baseline, the result at 256 dimensions with $\lambda = 2/3$ is reported in this case.

|  | MED | | CRAN | | CACM | | CISI | |
| --- | --- | --- | --- | --- | --- | --- | --- | --- |
|  | prec. | impr. | prec. | impr. | prec. | impr. | prec. | impr. |
| cos+tf | 44.3 | - | 29.9 | - | 17.9 | - | 12.7 | - |
| LSI | 51.7 | +16.7 | *28.7 | -4.0 | *16.0 | -11.6 | 12.8 | +0.8 |
| PLSI | 63.9 | +44.2 | 35.1 | +17.4 | 22.9 | +27.9 | 18.8 | +48.0 |
| PLSI* | 66.3 | +49.7 | 37.5 | +25.4 | 26.8 | +49.7 | 20.1 | +58.3 |

### 4.1 Perplexity Evaluation

In order to compare the predictive performance of PLSA and LSA one has to specify how to extract probabilities from a LSA decomposition. This problem is not trivial, since negative entries prohibit a simple re-normalization of the approximating matrix $\tilde{N}$. We have followed the approach of [2] to derive LSA probabilities.

Two data sets that have been used to evaluate the perplexity performance: (i) a standard information retrieval test collection MED with 1033 document, (ii) a dataset with noun-adjective pairs generated from a tagged version of the LOB corpus. In the first case, the goal was to predict word occurrences based on (parts of) the words in a document. In the second case, nouns have to predicted conditioned on an associated adjective. Figure 5 reports perplexity results for LSA and PLSA on the MED (a) and LOB (b) datasets in dependence on the number of dimensions of the (probabilistic) latent semantic space. PLSA outperforms the statistical model derived from standard LSA by far. On the MED collection PLSA reduces perplexity relative to the unigram baseline by more than a factor of three ($3073/936 \approx 3.3$), while LSA achieves less than a factor of two in reduction ($3073/1647 \approx 1.9$). On the less sparse LOB data the PLSA reduction in perplexity is $1316/547 \approx 2.41$ while the reduction achieved by LSA is only $1316/632 \approx 2.08$. In order to demonstrate the advantages of TEM, we have also trained aspect models on the MED data by standard EM with early stopping. As can be seen from the curves in Figure 5 (a), the difference between EM and TEM model fitting is significant. Although both strategies – tempering and early stopping – are successful in controlling the model complexity, EM training performs worse, since it makes a very inefficient use of the available degrees of freedom. Notice, that with both methods it is possible to train high-dimensional models with a continuous improvement in performance. The number of latent space dimensions may even exceed the rank of the co-occurrence matrix $\mathbf{N}$ and the choice of the number of dimensions becomes merely an issue of possible limitations of computational resources.

### 4.2 Information Retrieval

One of the key problems in information retrieval is *automatic indexing* which has its main application in query-based retrieval. The most popular family of information retrieval techniques is based on the *Vector Space Model* (VSM) for documents [12]. Here, we have utilized a rather straightforward representation based on the (untransformed) term frequencies $n(d, w)$ together with the standard cosine matching function, a more detailed experimental analysis can be found in [6]. The same representation applies to queries $q$, so that the matching function for the baseline term matching method can be written as

$$s(d, q) = \frac{\sum_w n(d, w) n(q, w)}{\sqrt{\sum_w n(d, w)^2} \sqrt{\sum_w n(q, w)^2}}, \quad (10)$$

In Latent Semantic Indexing (LSI), the original vector space representation of documents is replaced by a representation in the low-dimensional latent space and the similarity is computed based on that representation. Queries or documents which were not part of the original collection can be *folded in* by a simple matrix multiplication (cf. [3] for details). In our experiments, we have actually considered linear combinations of the original similarity score (10) (weight $\lambda$) and the one derived from the latent space representation (weight $1 - \lambda$).

The same ideas have been applied in Probabilistic Latent Semantic Indexing (PLSI) in conjunction with the PLSA model. More precisely, the low-dimensional representation in the *factor space* $P(z|d)$ and $P(z|q)$ have been utilized to evaluate similarities. To achieve this, queries have to be folded in, which is done in the PLSA by fixing the $P(w|z)$ parameters and calculating weights $P(z|q)$ by TEM.

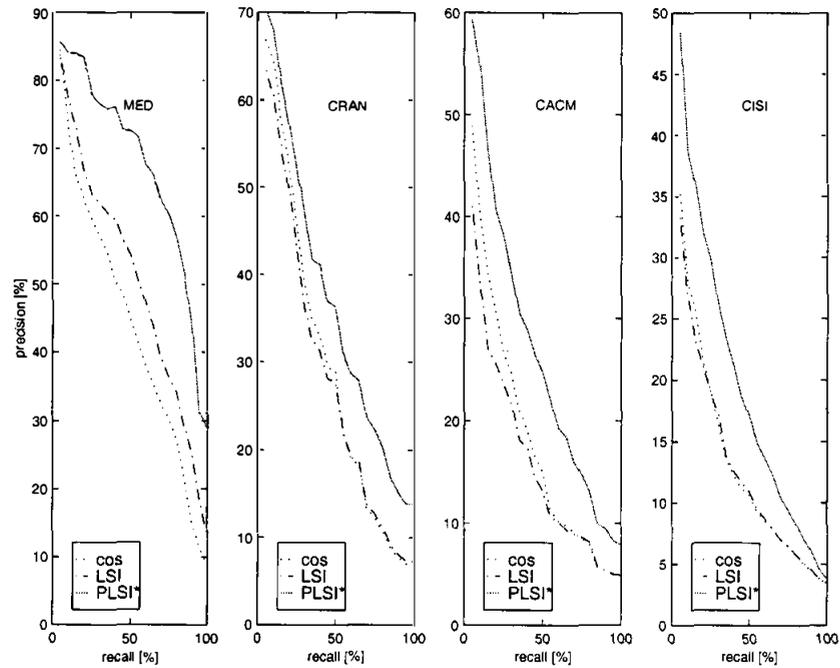

Figure 6: Precision-recall curves for term matching, LSI, and PLSI* on the 4 test collections.

One advantage of using statistical models vs. SVD techniques is that it allows us to systematically combine different models. While this should optimally be done according to a Bayesian model combination scheme, we have utilized a much simpler approach in our experiments which has nevertheless shown excellent performance and robustness. Namely, we have simply combined the cosine scores of all models with a uniform weight. The resulting method is referred to as PLSI*. Empirically we have found the performance to be very robust w.r.t. different (non-uniform) weights and also w.r.t. the $\lambda$-weight used in combination with the original cosine score. This is due to the noise reducing benefits of (model) averaging. Notice that LSA representations for different $K$ form a nested sequence, which is not true for the statistical models which are expected to capture a larger variety of reasonable decompositions.

We have utilized the following four medium-sized standard document collection with relevance assessment: (i) MED (1033 document abstracts from the National Library of Medicine), (ii) CRAN (1400 document abstracts on aeronautics from the Cranfield Institute of Technology), (iii) CACM (3204 abstracts from the CACM Journal), and (iv) CISI (1460 abstracts in library science from the Institute for Scientific Informa-tion). The condensed results in terms of average precision recall (at the 9 recall levels $10\% - 90\%$) are summarized in Table 1, while the corresponding precision recall curves can be found in Figure 6. Here are some additional details of the experimental setup: PLSA models at $K = 32, 48, 64, 80, 128$ have been trained by TEM for each data set with 10% held-out data. For PLSI we report the best result obtained by any of these models, for LSI we report the best result obtained for the optimal dimension (exploring 32–512 dimensions at a step size of 8). The combination weight $\lambda$ with the cosine baseline score has been coarsely optimized by hand, MED, CRAN: $\lambda = 1/2$, CACM, CISI:$\lambda = 2/3$.

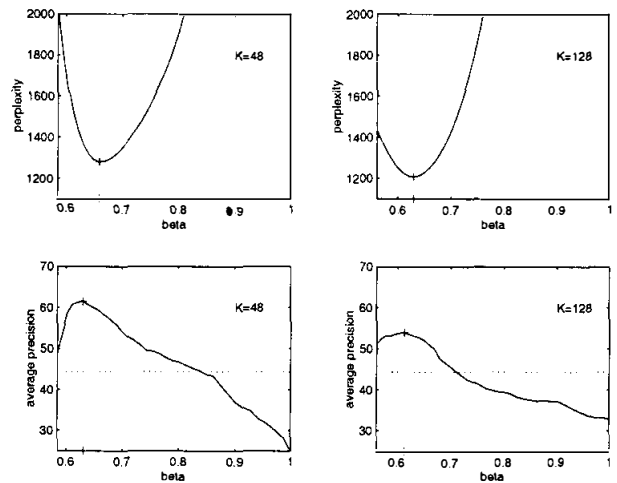

Figure 7: Perplexity and average precision as a function of the inverse temperature $\beta$ for an aspect model with $K = 48$ (left) and $K = 128$ (right).

The experiments consistently validate the advantages of PLSI over LSI. Substantial performance gains have been achieved for all 4 data sets. Notice that the relative precision gain compared to the baseline method is typically around 100% in the most interesting intermediate regime of recall! In particular, PLSI works well even in cases where LSI fails completely (these problems of LSI are in accordance with the original results reported in [3]). The benefits of model combination are also very substantial. In all cases the (uniformly) combined model performed better than the best single model. As a sight-effect model averaging also deliberated from selecting the correct model dimensionality.

These experiments demonstrate that the advantages of PLSA over standard LSA are not restricted to applications with performance criteria directly depending on the perplexity. Statistical objective functions like the perplexity (log-likelihood) may thus provide a general yardstick for analysis methods in text learning and information retrieval. To stress this point we ran an experiment on the MED data, where both, perplexity and average precision, have been monitored simultaneously as a function of $\beta$. The resulting curves which show a striking correlation are plotted in Figure 7.

## 5  Conclusion

We have proposed a novel method for unsupervised learning, called *Probabilistic Latent Semantic Analysis*, which is based on a statistical latent class model. We have argued that this approach is more principled than standard Latent Semantic Analysis, since it possesses a sound statistical foundation. *Tempered Expectation Maximization* has been presented as a powerful fitting procedure. We have experimentally verified the claimed advantages achieving substantial performance gains. Probabilistic Latent Semantic Analysis has thus to be considered as a promising novel unsupervised learning method with a wide range of applications in text learning and information retrieval.

**Acknowledgments**

The author would like to thank Jan Puzicha, Andrew McCallum, Mike Jordan, Joachim Buhmann, Tali Tishby, Nelson Morgan, Jerry Feldman, Dan Gildea, Andrew Ng, Sebastian Thrun, and Tom Mitchell for stimulating discussions and helpful hints. This work has been supported by a DAAD fellowship.